%% file: neurips_2025.tex
\documentclass{article}

\usepackage[numbers]{natbib}
\usepackage[preprint]{neurips_2026}

\usepackage[utf8]{inputenc} 
\usepackage[T1]{fontenc}    
\usepackage{hyperref}       
\usepackage{url}            
\usepackage{booktabs}       
\usepackage{amsfonts}       
\usepackage{nicefrac}       
\usepackage{microtype}      
\usepackage{xcolor}         

\usepackage{graphicx}
\usepackage{booktabs}
\usepackage{animate}
\usepackage{caption}

\usepackage{tcolorbox}

\tcbuselibrary{listings, breakable} 

\title{OmniHumanoid: Streaming Cross-Embodiment Video Generation with Paired-Free Adaptation}

\author{
  Yiren Song \quad
  Xiyao Deng \quad
  Pei Yang \quad
  Yihan Wang \quad
  Mike Zheng Shou\thanks{Corresponding author.} \\
  \\
  Show Lab, National University of Singapore
}

\begin{document}

\maketitle


\vspace{-10mm}

\begin{figure}[htbp]
\centering
\includegraphics[width=\linewidth]{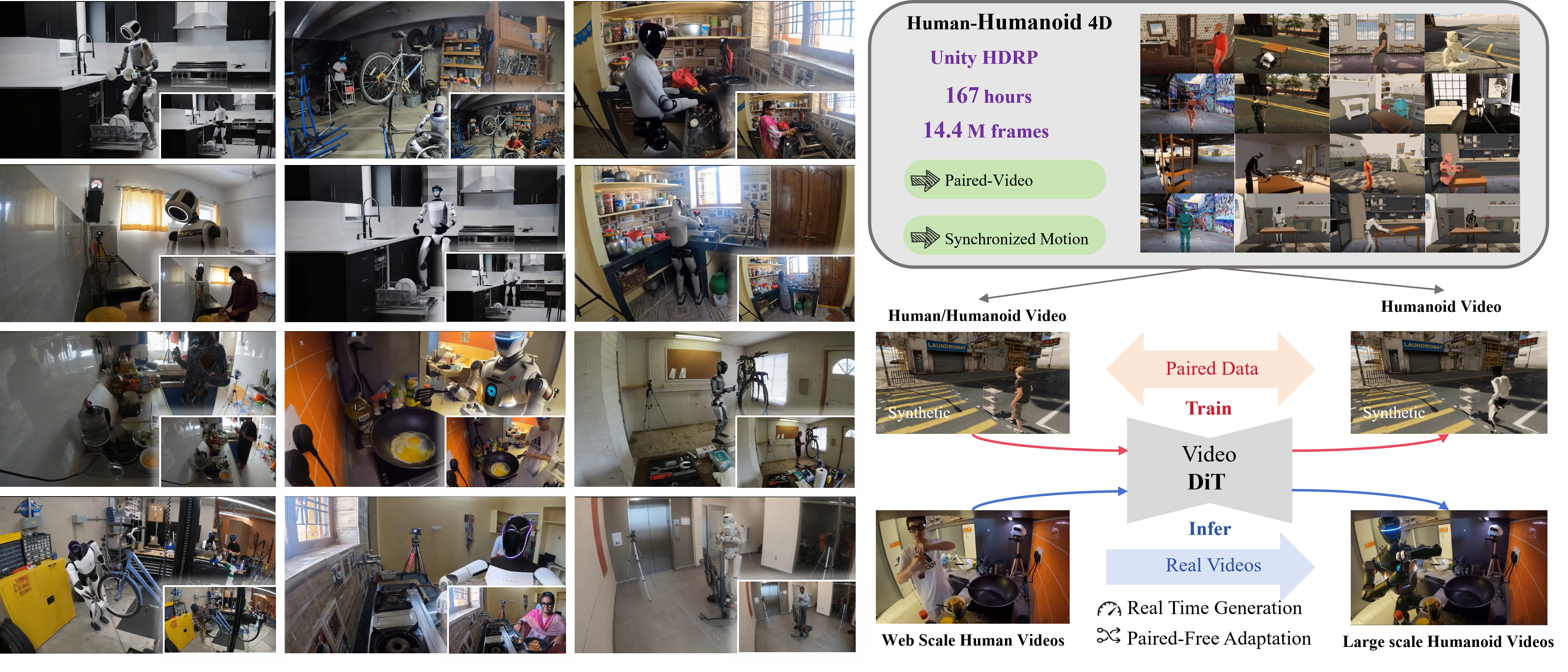}
\vspace{-5mm}
\caption{OmniHumanoid enables scalable cross-embodiment video generation by decoupling transferable motion from embodiment-specific appearance. Built on a motion-aligned synthetic dataset spanning diverse humanoids, scenes, and actions, our framework transfers motion to new embodiments and synthesizes consistent target videos with only lightweight adaptation.}
\label{fig:teaser}
\end{figure}

\begin{abstract}

Cross-embodiment video generation aims to transfer motions across different humanoid embodiments, such as human-to-robot and robot-to-robot, enabling scalable data generation for embodied intelligence. A major challenge in this setting is that motion dynamics are partly transferable across embodiments, whereas appearance and morphology remain embodiment-specific. Existing approaches often entangle these factors, and many require paired data for every target embodiment, which limits scalability to new robots. We present OmniHumanoid, a framework that factorizes transferable motion learning and embodiment-specific adaptation. Our method learns a shared motion transfer model from motion-aligned paired videos spanning multiple embodiments, while adapting to a new embodiment using only unpaired videos through lightweight embodiment-specific adapters. To reduce interference between motion transfer and embodiment adaptation, we further introduce a branch-isolated attention design that separates motion conditioning from embodiment-specific modulation. In addition, we construct a synthetic cross-embodiment dataset with motion-aligned paired videos rendered across diverse humanoid assets, scenes, and viewpoints. Experiments on both synthetic and real-world benchmarks show that OmniHumanoid achieves strong motion fidelity and embodiment consistency, while enabling scalable adaptation to unseen humanoid embodiments without retraining the shared motion model.  Code is released at \href{https://github.com/showlab/OmniHumanoid}{https://github.com/showlab/OmniHumanoid}

\end{abstract}

\input{section/introduction.tex}

\input{section/method.tex}
\input{section/experiment.tex}
\input{section/conclusion.tex}

\bibliographystyle{plainnat} 
\bibliography{main}

\newpage
\appendix

\setcounter{page}{1}

\setcounter{section}{0}
\renewcommand{\thesection}{\Alph{section}}


\begin{center}
\Large\textbf{Appendix}
\end{center}


\section{Prompt for Automated Evaluation}

We employ a comprehensive prompt structure to evaluate the generated videos across four distinct dimensions. The complete prompt template is shown below:

\begin{tcblisting}{colback=gray!5, colframe=gray!40, arc=4pt, title=\textbf{Comprehensive Evaluation Prompt}, listing only, breakable, listing options={basicstyle=\ttfamily\small, breaklines=true, columns=fullflexible}}
# =========================
# PROMPT: MOTION FIDELITY
# =========================
You are an expert evaluator for video generation and motion tracking.

You will see TWO videos:
(1) Condition video (Human): the original reference clip showing the driving motion.
(2) Generated video (Robot): the synthesized output.

TASK: Evaluate MOTION FIDELITY of the generated robot compared to the human condition video.
Focus on:
- Movement accuracy: Does the generated robot's motion (pose, trajectory, speed, and timing) precisely match the human's motion?
- Articulation: Are the joint movements, limb articulations, and subtle actions faithful to the human reference?
- Smoothness: Is the motion temporally smooth without unnatural jitter or sudden jumps?

SCORING (1-10):
10: Motion perfectly matches the condition video; fluid and flawless.
8-9: Very accurate motion; only extremely minor differences in timing or subtle jitter.
6-7: Noticeable differences in motion or occasional jitter, but the general action matches.
4-5: Frequent motion errors, stiffness, or desynchronization.
1-3: Motion completely breaks down or fails to follow the condition video.

OUTPUT FORMAT:
Return ONLY a valid JSON object exactly like:
{"score": <number 1-10>, "reason": "<one short sentence with the main evidence>"}

# =========================
# PROMPT: EMBODIMENT SIMILARITY
# =========================
You are an expert evaluator for identity and embodiment preservation in video synthesis.

You will see TWO files:
(1) Reference Image (Robot): The ground-truth appearance of the target robot.
(2) Generated Video (Robot): The synthesized output video.

TASK: Evaluate EMBODIMENT SIMILARITY  of the generated robot compared to the reference image.
Focus on:
- Identity preservation: Does the robot in the video look EXACTLY like the robot in the reference image?
- Structural integrity: Are body proportions, mechanical features, colors, and structural details faithfully preserved?
- Temporal consistency: Does the robot morph, warp, or change shape/identity across frames in the video?

SCORING (1-10):
10: Perfect preservation of identity, proportions, and appearance; zero morphing.
8-9: Very high similarity; minor softening or tiny structural shifts.
6-7: Noticeable morphing or differences in details, but still clearly the same robot.
4-5: Severe morphing; proportions feel wrong; identity is barely maintained.
1-3: Robot is unrecognizable or structurally completely broken compared to the image.

OUTPUT FORMAT:
Return ONLY a valid JSON object exactly like:
{"score": <number 1-10>, "reason": "<one short sentence with the main evidence>"}

# =========================
# PROMPT: BACKGROUND CONSISTENCY
# =========================
You are an expert evaluator for video background stability.

You will see TWO videos:
(1) Condition video (Human): the original reference clip.
(2) Generated video (Robot): the synthesized output.

TASK: Evaluate BACKGROUND CONSISTENCY of the generated video.
Focus on:
- Stability: Does the static background remain completely rigid and still over time?
- Fidelity: Are the background details, textures, and structures preserved exactly as in the condition video?
- Artifacts: Check for background flickering, texture crawling, unwanted warping near the moving robot, or "hallucinated" elements.

SCORING (1-10):
10: Background is perfectly stable, artifact-free, and identical to the condition video.
8-9: Mostly stable; very minor flickering or tiny artifacts in complex areas.
6-7: Noticeable background warping or flickering, especially near the moving robot.
4-5: Frequent background distortion, crawling textures, or loss of detail.
1-3: Background is extremely unstable, totally warped, or completely hallucinated.

OUTPUT FORMAT:
Return ONLY a valid JSON object exactly like:
{"score": <number 1-10>, "reason": "<one short sentence with the main evidence>"}

# =========================
# PROMPT: OVERALL QUALITY
# =========================
You are an expert evaluator for cinematic video quality.

You will see ONE video:
(1) Generated video (Robot): the synthesized output.

TASK: Evaluate OVERALL QUALITY of the generated video.
Focus on:
- Visual fidelity: Sharpness, clarity, and lack of noise/blur.
- Realism: Does the lighting, rendering, and compositing look professional and aesthetically pleasing?
- Artifact absence: Are there any glaring AI generation artifacts (e.g., weird pixelation, smearing, color shifts)?
- Holistic impression: How pleasing and professional is the final result?

SCORING (1-10):
10: Flawless cinematic quality; hyper-realistic and sharp.
8-9: High quality; very clean with only barely noticeable AI artifacts.
6-7: Acceptable quality; some visible noise, blur, or slight AI artifacts, but highly watchable.
4-5: Poor quality; obvious generation artifacts, severe blur, or color degradation.
1-3: Unwatchable; completely destroyed by artifacts or extreme blur.

OUTPUT FORMAT:
Return ONLY a valid JSON object exactly like:
{"score": <number 1-10>, "reason": "<one short sentence with the main evidence>"}
\end{tcblisting}

\section{User Study Results}

We conducted a user study to evaluate the generated videos across four key dimensions: Motion Fidelity, Embodiment Similarity, Background (BG) Consistency, and Overall Quality. Participants were asked to vote for the best method in each category. The results, representing the percentage of total votes each method received, are summarized in Table~\ref{tab:user_study}. As shown, our method significantly outperforms existing baselines across all metrics.

\begin{table}[htbp]
\centering
\caption{User study results. Numbers indicate the percentage of user preference votes each method received across four evaluation metrics.}
\label{tab:user_study}

\begin{tabular}{l c c c c}
\toprule
Method & Motion Fidelity & Embodiment Similarity & BG Consistency & Overall Quality \\
\midrule
Kling O1 & 7.1\% & 14.1\% & 11.1\% & 8.1\% \\
Kling O3 & 12.1\% & 16.2\% & 17.2\% & 20.2\% \\
\textbf{Ours} & \textbf{72.7\%} & \textbf{65.7\%} & \textbf{62.6\%} & \textbf{63.6\%} \\
Runway Aleph & 8.1\% & 4.0\% & 9.1\% & 8.1\% \\
\bottomrule
\end{tabular}
\end{table}

\section{Limitations}
\label{sec:limitations}

While our proposed method significantly improves generation efficiency, it still possesses certain limitations. Specifically, we observe a noticeable performance degradation in the few-step distillation regime compared to the original teacher model. When operating at extremely low inference steps (e.g.4 steps), the generated videos may exhibit slight compromises in fine-grained details, temporal smoothness, or complex motion fidelity when directly compared to the high-quality outputs generated by the full-step teacher model. This phenomenon highlights an inherent trade-off between inference speed and generation quality within current distillation frameworks. The distilled student model, although highly efficient in reducing computational overhead, still struggles to perfectly approximate the complex data distribution captured by the teacher model over dozens of denoising steps. In future work, we plan to explore more advanced knowledge distillation strategies to mitigate this issue. 


\end{document}

%% file: section/introduction.tex
\section{Introduction}
\vspace{-3mm}

Learning from videos has become a promising paradigm for embodied intelligence, as videos naturally capture rich motion, interaction, and scene dynamics. However, collecting high-quality demonstrations for every robot embodiment is expensive and difficult to scale. A more scalable alternative is to transform existing demonstrations, such as human videos or videos of other robots, into videos of a target robot embodiment, and use the generated data for downstream robot learning. This motivates cross-embodiment video generation: given a source video and a target humanoid embodiment, synthesize a target video that preserves the source motion while rendering the desired robot appearance and morphology.

Despite its potential, cross-embodiment video generation remains far from solved. Existing methods face four major obstacles: motion is often entangled with embodiment-specific geometry and kinematics; paired source-target videos are expensive to collect for every new robot; generic video editing models struggle to preserve robot identity, structural details, and temporal consistency for high-DOF robotic bodies; and high-quality video-to-video generation is typically too slow for interactive use or large-scale data production. These limitations make it difficult to build a cross-embodiment generator that is both scalable and practically deployable.

To address these issues, we propose \textbf{OmniHumanoid} under the \textbf{TAPE} principle: \textbf{T}ransferable motion, paired-free \textbf{A}daptation, embodiment \textbf{P}reservation, and generation \textbf{E}fficiency. OmniHumanoid decomposes cross-embodiment generation into two complementary components: a shared motion component capturing how actions and scenes evolve, and an embodiment-specific appearance component determining the target humanoid's look and movement. This design allows motion knowledge to be reused across humanoids, while new robots are introduced through lightweight unpaired adaptation. 

To realize this factorization, OmniHumanoid introduces a unidirectional information-flow design separating motion understanding from embodiment rendering. The model learns transferable motion from paired multi-embodiment videos, while robot-specific appearance and morphology are captured by lightweight embodiment adapters trained from unpaired videos. By preventing embodiment-specific priors from entering the shared motion pathway, OmniHumanoid preserves general motion transfer ability while allowing efficient adaptation to unseen humanoids.

To support this learning paradigm, we construct a motion-aligned synthetic cross-embodiment dataset. Diverse motions are retargeted to multiple digital humans and robots under aligned scenes, cameras, and temporal structures. The resulting videos share underlying motion and scene dynamics while differing in appearance and morphology, providing controlled supervision for motion transfer and systematic evaluation on unseen robots, actions, and environments.

Beyond paired-free adaptation, OmniHumanoid targets efficient deployment. We introduce a streaming video-to-video distillation pipeline turning the expensive bidirectional generator into an efficient causal generator. This reduces generation from 50 denoising steps to 4 and enables autoregressive cross-embodiment video generation at 720p and 5 FPS on a single NVIDIA H200 GPU.

Extensive experiments on synthetic and real-world benchmarks show that OmniHumanoid achieves strong motion fidelity and embodiment consistency, supports paired-free adaptation to unseen humanoids, and enables efficient long-horizon generation. Compared with generic video editing and offline cross-embodiment pipelines, OmniHumanoid offers a more scalable and practical path toward robot-centric video data generation.

Our contributions are threefold:
\begin{itemize}
    \item We propose \textbf{OmniHumanoid}, a factorized cross-embodiment video generation framework built upon the \textbf{TAPE} principle. Through a unidirectional information-flow design, OmniHumanoid decouples transferable motion from embodiment-specific appearance, enabling paired-free adaptation while preserving target embodiment identity.

    \item We introduce a streaming video-to-video distillation pipeline that converts the bidirectional generator into a causal attention-based model, enabling efficient long-horizon and real-time cross-embodiment video generation.

    \item We construct a motion-aligned synthetic cross-embodiment dataset across diverse humanoids, motions, scenes, and viewpoints, and validate generalization to unseen embodiments, actions, and environments on both synthetic and real-world benchmarks.
\end{itemize}

\section{Related Work}
\vspace{-2mm}

\subsection{Video Generation Models}
\vspace{-2mm}

In recent years, diffusion models have been widely applied to image synthesis \cite{rombach2022high}, image editing \cite{brooks2023instructpix2pix, hertz2022prompt, huang2025photodoodle}, video generation \cite{blattmann2023stable, song2025worldwander, chen2025transanimate, guo2023animatediff, ma2024followyouremoji, ma2024followpose, ma2025followyourclick, ma2025followyourmotion, li2025ic}, and procedural generation \cite{song2024processpainter, song2025layertracer, song2025makeanything, ye2025loom}. Within video generation, the currently prevalent DiT architecture \cite{ho2020denoising} has progressively surpassed earlier GAN-based \cite{pan2017create} and UNet-based methods \cite{guo2023animatediff, song2024processpainter} by significantly enhancing visual fidelity and temporal consistency. Today, rapidly evolving DiT frameworks form the foundation of cutting-edge models like WAN \cite{wan2025wan}, Sora \cite{liu2024sora}, and HunyuanVideo \cite{kong2024hunyuanvideo}, pushing generation quality to new heights. Recent studies have further leveraged large pre-trained video generation models for robotic and manipulation tasks \cite{fu2025learning, ci2025h2r, yuan2026fast, kim2026cosmos, li2025unified}, highlighting their potential for cross-domain generalization and interactive learning.

\subsection{Robotizing Human Videos}
\vspace{-2mm}

In embodied intelligence, the scarcity of large-scale robot image and video datasets remains a major bottleneck, motivating recent efforts to bridge the embodiment gap by converting human-centric videos into robot-centric observations. In first-person settings, methods jointly train on egocentric human and robot data \cite{yang2025egovla, kareer2025egomimic, qiu2025humanoid}, where EgoVLA \cite{yang2025egovla} trains vision-language models on both domains, and EgoMimic \cite{kareer2025egomimic} and PH2D \cite{qiu2025humanoid} align wearable-camera human demonstrations with teleoperated robot trajectories through shared backbones and cross-domain losses. Data-editing pipelines further scale supervision by converting in-the-wild egocentric human videos into robot-centric observations \cite{lepert2025masquerade}. Extending robotization to third-person videos is more challenging due to full-body motion, dynamic backgrounds, complex interactions, and self-occlusion. Phantom \cite{lepert2025phantom} overlays rendered robotic arms guided by estimated hand poses, but rendering-based pipelines are limited by lighting, depth, scene geometry, and generalization to unseen robot embodiments. Recent diffusion-based human-to-robot video translation methods address third-person robotization with full-body motion and occlusion handling beyond simple overlays \cite{yang2025x, song2025mitty}. Building on this direction, we propose OmniHumanoid, a modularly decoupled framework that learns general motion representations with a shared motion model and uses enhanced embodiment LoRA modules for robot appearance replacement under structural guidance, enabling robust human-to-robot video translation in complex motions and cluttered scenes.

\subsection{Decoupled Learning}
\vspace{-3mm}
Decoupled learning has become an effective strategy for improving the generalization and controllability of generative visual models by separating factors of variation with different statistical structures. In video generation, AnimateDiff \cite{guo2023animatediff} decouples spatial and temporal modeling with lightweight Temporal Modules, enabling image generation models to propagate information across time. MotionDirector \cite{zhao2024motiondirector} uses dual-path LoRA to disentangle motion and appearance, separating “how to move” from “what to look like” for better transfer across appearances. In image stylization, many methods decouple style and content to balance stylistic consistency and content preservation \cite{wang2024instantstyle, xing2024csgo}. OmniConsistency \cite{song2025omniconsistency} further reduces the conflict between style learning and consistency preservation through a two-stage progressive LoRA design. Overall, these methods reduce interference between different factors, leading to improved generation quality and stronger generalization.

%% file: section/method.tex
\section{Methods}
\label{sec:method}
\vspace{-4mm}
We present \textbf{OmniHumanoid}, a scalable and streaming framework for cross-embodiment video generation built around the TAPE principle. Sec.~\ref{sec:problem} defines paired training and paired-free adaptation; Sec.~\ref{sec:architecture} introduces our branch-decoupled architecture; Sec.~\ref{sec:training} describes two-stage training and unseen-embodiment adaptation; and Sec.~\ref{sec:streaming} presents streaming video-to-video distillation for efficient long-horizon inference.

\subsection{Problem Formulation}
\label{sec:problem}
\vspace{-3mm}
Given a source video $V^{\mathrm{src}}$ and a target embodiment $e$, cross-embodiment video generation aims to synthesize a target video $\hat{V}^{e}$ that preserves the motion and scene dynamics of $V^{\mathrm{src}}$ while rendering the appearance and morphology of embodiment $e$.

During training, we assume access to a motion-aligned paired dataset:
\begin{equation}
\mathcal{D}_{\mathrm{pair}}
=
\{(V_i^{\mathrm{src}}, V_i^{\mathrm{tgt}}, e_i)\}_{i=1}^{N},
\end{equation}
where $V_i^{\mathrm{src}}$ is a source video, $V_i^{\mathrm{tgt}}$ is the corresponding target video under embodiment $e_i$, and the paired videos share the same underlying motion and scene dynamics. These paired samples provide supervision for learning motion transfer across embodiments. At adaptation time, for a new embodiment $e^\ast$, paired source--target videos are unavailable. Instead, we only assume access to unpaired videos:
\vspace{-0.3em}
\begin{equation}
\mathcal{U}_{e^\ast}=\{V_j^{e^\ast}\}_{j=1}^{M}.
\end{equation}
\vspace{-0.3em}
The goal is to adapt the model to $e^\ast$ using $\mathcal{U}_{e^\ast}$ only, without retraining the shared motion transfer parameters. This setting motivates a factorized design that separates transferable motion learning from embodiment-specific appearance and morphology modeling.

\begin{figure*}[tbp]
    \centering
    \includegraphics[width=\textwidth]{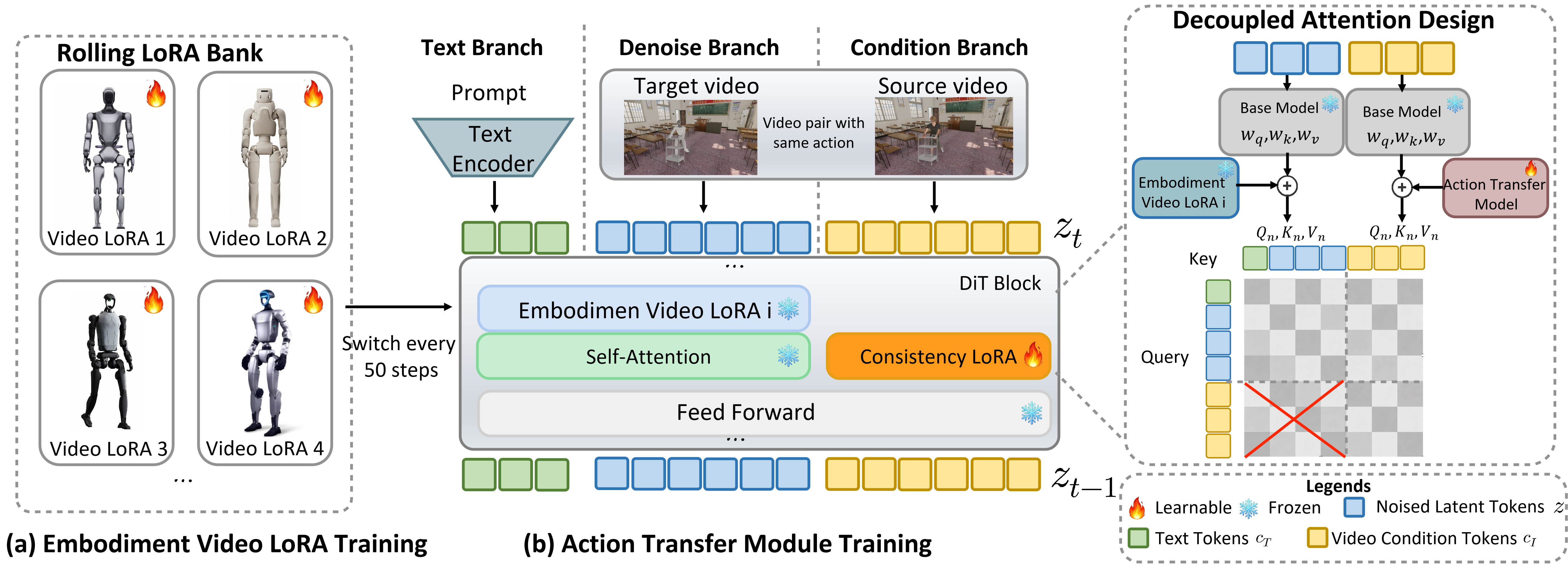}
    \vspace{-5mm}
    \caption{Overview of OmniHumanoid. A Shared Motion Transfer Model learns transferable motion from motion-aligned paired videos, while each Embodiment Video LoRA captures the appearance and morphology of a specific humanoid embodiment from unpaired videos. During paired training, the target embodiment LoRA is activated on the denoising branch, while the shared motion transfer parameters are optimized on the conditioning branch. This branch-decoupled design enables paired-free adaptation to unseen embodiments while preserving transferable motion learning.}
    \label{fig:method_overview}
\end{figure*}

\subsection{Architecture and Unidirectional Motion-Appearance Decoupling}
\label{sec:architecture}

OmniHumanoid is built on a Diffusion Transformer (DiT) video generation backbone. The central design is to decompose cross-embodiment generation into two complementary components: a \textbf{Shared Motion Transfer Model} that captures embodiment-invariant motion dynamics, and a bank of \textbf{Embodiment Video LoRAs} that model embodiment-specific appearance and morphology.

\paragraph{Shared Motion Transfer Model.}
The shared motion transfer model operates on the conditioning branch. It takes source video $V^{\mathrm{src}}$ to encode transferable motion cues like temporal dynamics, object interactions, and scene evolution. Being shared across all embodiments, this module learns common humanoid behaviors: how actions unfold over time.

\paragraph{Embodiment Video LoRA.}
For each embodiment $e$, we introduce a lightweight LoRA module $\Phi_e$ attached to the denoising branch. $\Phi_e$ captures embodiment-specific geometry, texture, and structural priors. At inference time, the target video is generated by loading the corresponding embodiment LoRA while keeping the shared motion transfer model fixed.

\paragraph{Branch-specific parameterization.}
We denote the token streams in the DiT backbone by
\[
X^{\mathrm{text}}, \qquad X^{\mathrm{den}}, \qquad X^{\mathrm{cond}},
\]
corresponding to text tokens, denoising latent tokens, and source-video conditioning tokens. For branch $b\in\{\mathrm{text},\mathrm{den},\mathrm{cond}\}$, the standard attention projections are:
\vspace{-0.3em}
\begin{equation}
Q^b=X^bW_Q,\qquad
K^b=X^bW_K,\qquad
V^b=X^bW_V.
\end{equation}
\vspace{-0.3em}

For a target embodiment $e$, we apply LoRA updates only to the denoising branch:
\begin{equation}
W_{\bullet}^{\mathrm{den},e}
=
W_{\bullet}
+
\Delta W_{\bullet}^{(e)},
\qquad
\Delta W_{\bullet}^{(e)}
=
B_{\bullet}^{(e)}A_{\bullet}^{(e)},
\qquad
\bullet\in\{Q,K,V\},
\end{equation}
where $A_{\bullet}^{(e)}\in\mathbb{R}^{r\times d}$, $B_{\bullet}^{(e)}\in\mathbb{R}^{d\times r}$, and $r\ll d$. The text and conditioning branches use the shared backbone projections without embodiment-specific updates.

\paragraph{Unidirectional information flow.}
The key design is not only where the embodiment LoRA is attached, but also how information is allowed to flow. To prevent embodiment-specific appearance priors from directly affecting the shared motion pathway, we apply an asymmetric attention mask:
\begin{equation}
\mathcal{M}(\mathrm{den}\rightarrow\mathrm{cond}) = 1,
\qquad
\mathcal{M}(\mathrm{cond}\rightarrow\mathrm{den}) = 0,
\end{equation}
where $1$ and $0$ denote visible and masked attention, respectively. Thus, the denoising branch can read motion information from the conditioning branch, while the conditioning branch is isolated from embodiment-specific denoising updates. This unidirectional information flow helps preserve transferable motion representations while allowing the denoising branch to render target-specific morphology and appearance.

\subsection{Two-Stage Training and Paired-Free Adaptation}
\label{sec:training}

The factorized architecture naturally leads to a two-stage training strategy. The first stage teaches each embodiment LoRA what a target embodiment looks like, while the second stage teaches the shared model how motion transfers across embodiments.

\paragraph{Stage I: Embodiment Video LoRA Pretraining.}
For each seen embodiment $e$, we train $\Phi_e$ using only unpaired videos $\mathcal{U}_e$. The DiT backbone is frozen, and only the LoRA parameters are optimized under the standard video diffusion denoising objective. Since $\Phi_e$ is attached to the denoising branch, it learns embodiment-specific appearance, geometry, and morphology directly in the generative pathway without requiring paired motion supervision.

\paragraph{Stage II: Shared Motion Transfer Training.}
We then train the Shared Motion Transfer Model on the paired dataset $\mathcal{D}_{\mathrm{pair}}$ while freezing all Embodiment Video LoRAs. For each paired sample $(V^{\mathrm{src}}, V^{\mathrm{tgt}}, e)$, we activate the LoRA $\Phi_e$ corresponding to the target embodiment and optimize only the shared motion transfer parameters. To reduce bias toward any single embodiment, we adopt a \emph{rolling LoRA loading} strategy: the active LoRA changes across training iterations according to the target embodiment of the current batch. This encourages the shared model to focus on cross-embodiment invariances, namely motion and scene dynamics, while leaving embodiment-specific rendering to the LoRAs.

\paragraph{Adaptation to Unseen Embodiments.}
For a new humanoid embodiment $e^\ast$, we instantiate a new LoRA $\Phi_{e^\ast}$ and train it using only a small collection of unpaired videos $\mathcal{U}_{e^\ast}$, typically on the order of tens of videos, without requiring any motion-aligned source--target pairs. The Shared Motion Transfer Model is kept frozen throughout this adaptation stage. During inference, OmniHumanoid takes a source video $V^{\mathrm{src}}$ and the target embodiment $e^\ast$, loads the adapted LoRA $\Phi_{e^\ast}$, and synthesizes a target video that transfers the source motion to the new robot embodiment.

\subsection{Streaming Video-to-Video Distillation}
\label{sec:streaming}

While the decoupled bidirectional generator provides high-quality cross-embodiment synthesis, standard DiT-based generation requires iterative full-sequence denoising, making it computationally expensive for interactive use and large-scale data production. To improve efficiency, we distill the bidirectional OmniHumanoid generator into a causal streaming student.

We organize the reference, source-video condition, and target-video tokens into an interleaved super-chunk layout:
\begin{equation}
[
\mathrm{ref}
\mid
\mathrm{cond}_0
\mid
\mathrm{tgt}_0
\mid
\cdots
\mid
\mathrm{cond}_M
\mid
\mathrm{tgt}_M
].
\end{equation}
During inference, target chunk $i$ attends only to the reference, previous chunks, and the current condition chunk, while future chunks are masked out. This block-wise causal structure enables autoregressive rollout with cached key-value states.

We train the causal student with a two-stage distillation objective. First, the student is initialized from the bidirectional teacher and optimized with teacher-forcing denoising score matching under the causal mask, which preserves basic token prediction fidelity. We then apply self-forcing few-step distillation, where the student is optimized under its own autoregressive rollout trajectory with teacher-guided distribution matching:
\begin{equation}
\mathcal{L}_{\mathrm{stream}}
=
\mathcal{L}_{\mathrm{DSM}}
+
\lambda_{\mathrm{vsd}}\mathcal{L}_{\mathrm{VSD}}
+
\lambda_{\mathrm{gan}}\mathcal{L}_{\mathrm{GAN}}.
\end{equation}
Here, $\mathcal{L}_{\mathrm{DSM}}$ denotes the denoising score matching loss under causal attention, $\mathcal{L}_{\mathrm{VSD}}$ aligns the fast causal student with the frozen bidirectional teacher's score guidance, and $\mathcal{L}_{\mathrm{GAN}}$ improves local sharpness and reduces detail degradation in few-step generation. This distillation converts the 50-step bidirectional generator into a 4-step causal streaming model, enabling efficient long-horizon cross-embodiment video generation.

\section{Synthetic Cross-Embodiment Dataset}
\label{sec:dataset}

To support the factorized learning paradigm and evaluate cross-embodiment generalization, we construct a motion-aligned synthetic video dataset with a Unity-based rendering pipeline. The dataset is built on the Humoto motion library, which provides over 700 humanoid motion sequences covering object manipulation, environment interaction, locomotion, and daily full-body activities.

\begin{figure*}[tbp]
    \centering
    \includegraphics[width=1.0\textwidth]{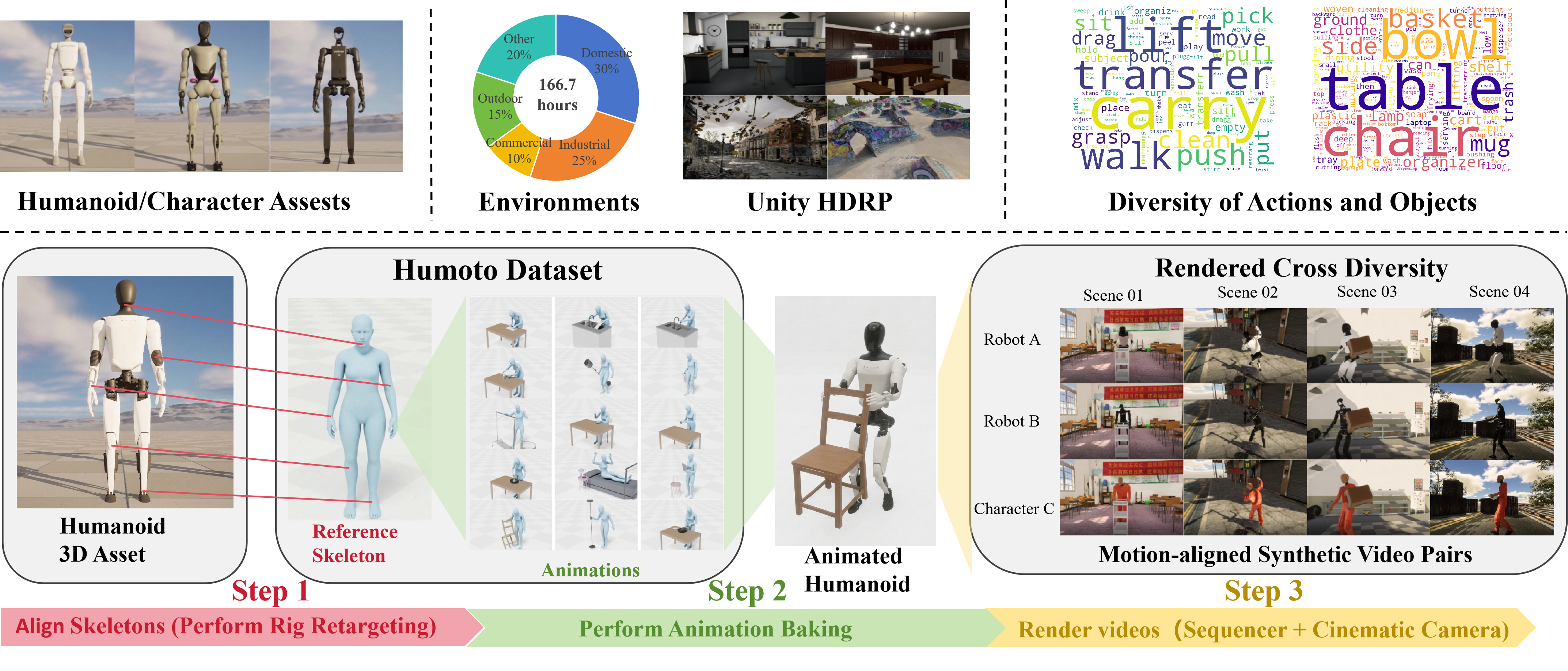}
    \vspace{-7mm}
    \caption{Synthetic data construction pipeline. We create paired human-humanoid videos from community character, animation, and environment assets by aligning skeletons, retargeting the same motions to different embodiments, and recording them in diverse scenes with identical camera setups.}
    \label{fig:data_pipeline}
\end{figure*}

\paragraph{Motion-aligned rendering.}
We curate ten humanoid assets, including five humanoid robots and five digital human avatars. To ensure reliable cross-embodiment motion alignment, all assets are equipped with compatible skeletons and animation controllers. We align their skeletons to a unified topology in Blender and perform animation retargeting in Unity, establishing joint-level correspondence across embodiments under a shared motion source.

\paragraph{Scene and embodiment variation.}
Each selected motion sequence is retargeted to all training embodiments. We render videos in 100 diverse scenes from online 3D platforms, covering offices, factories, and outdoor spaces. Within each scene, we vary only the humanoid asset while keeping the scene layout, camera viewpoint, and motion sequence unchanged. Thus, paired videos share identical motion and context, differing primarily in embodiment appearance and morphology. All videos are rendered at $1920\times1080$ resolution and 30 FPS, each containing roughly 300 frames.

\paragraph{Paired sample construction and evaluation split.}
To construct paired supervision, we form ordered source--target pairs among the training embodiments under matched motion and scene configurations, resulting in 7{,}200 paired training samples across 100 scenes. To evaluate generalization to unseen embodiments, one robot embodiment is fully held out from all training stages and used exclusively for testing. The test set also includes unseen motion tasks and scene configurations, ensuring that evaluation reflects cross-embodiment, cross-motion, and cross-environment generalization rather than memorization of seen embodiment--motion combinations.

%% file: section/experiment.tex
\begin{figure}[t] \centering \animategraphics[width=0.95\linewidth]{5}{section/result_fig/frame_}{00}{09} 
\caption{
Generalization to new scenes, tasks, and both seen and unseen robot embodiments. The top five rows show embodiments seen during training, while the bottom five show unseen ones. Given a reference video and a target humanoid-embodiment LoRA, our model transfers the motion to the target robot under novel environments and action configurations absent from paired supervision. The generated videos preserve action consistency, embodiment identity, and scene-consistent relighting. Readers can click and play the clips in this figure using {\color{red}\textbf{Adobe Acrobat}}.
}
\vspace{-4mm}
\label{fig:teaser} \end{figure}

\begin{figure}[t] \centering \animategraphics[width=0.95\linewidth]{5}{section/compare_fig/frame_}{00}{09} 
\vspace{-2mm}
\caption{
Qualitative comparison of motion transfer on various robot embodiments. Compared to general video generation baselines, our method generates significantly higher fidelity motions with fewer visual artifacts.
Our framework enables transferable motion synthesis across arbitrary humanoid types without retraining the core model, achieving embodiment-consistent generation at 720P resolution. Readers can click and play the video clips in this figure using {\color{red}\textbf{Adobe Acrobat}}.
}
\label{fig:compare} \end{figure}

\section{Experiments}
\vspace{-3mm}

\subsection{Experimental Setup}
\vspace{-2mm}

\noindent\textbf{Implementation Details.}
We finetune our model based on the Wan2.2-TI2V-5B backbone using rank-64 LoRA adaptation.
Training is performed for 10000 optimization steps on 2 NVIDIA H200 GPUs, with a batch size of 1 per GPU.
We adopt the AdamW optimizer with a learning rate of $2e-4$.
More implementation details are provided in the supplementary material.

\paragraph{Baselines.}
We compare with state-of-the-art V2V generation and editing systems, including commercial APIs such as Runway Gen-4 \cite{runway2025gen4}, Kling O1 \cite{team2025kling}, and Kling O3 \cite{kling2026o3}, as well as open-source methods such as Wan2.1-VACE \cite{wan2025wan} and the closely related X-Humanoid \cite{yang2025x}. Commercial APIs are evaluated as zero-shot reference-guided models without embodiment-specific adaptation, while trainable open-source baselines, including Wan2.1-VACE \cite{wan2025wan} and X-Humanoid \cite{yang2025x}, are fine-tuned on the same synthetic dataset following their original training strategies.

\paragraph{Benchmarks.}
We evaluate OmniHumanoid on two complementary benchmarks. The first is the Synthetic Held-out Embodiment Benchmark, where we reserve one unseen robot model, Unitree-G1, from training and use 50 motion-aligned videos for testing. Since ground-truth target videos are available in this synthetic setting, it supports reference-based quantitative evaluation. The second is the Real-world Benchmark, which contains 50 in-the-wild videos collected from human daily activities and online robot demonstrations. These videos cover diverse indoor tasks and scenes such as kitchens, garages, laboratories, and theaters, enabling reference-free evaluation under realistic visual conditions.

\paragraph{Metrics.}
We evaluate OmniHumanoid under two benchmark settings: the Synthetic Held-out Embodiment Benchmark with ground-truth target videos, and the Real-world Benchmark without paired ground truth. For the synthetic benchmark, we report PSNR, SSIM, and MSE to measure pixel-level reconstruction quality and motion-aligned visual fidelity between generated and ground-truth robot videos. In addition, for both benchmarks, we use Gemini-3 Flash as a vision-language evaluator to assess reference-free consistency along three dimensions: background consistency between the input and generated videos, motion consistency between the source and generated robot motions, and embodiment identity consistency reflecting whether the generated video maintains coherent robot-specific appearance. Thus, the synthetic benchmark supports both reference-based reconstruction metrics and reference-free consistency metrics, while the real-world benchmark is evaluated only with reference-free metrics.

\subsection{Comparison and Evaluation}
\vspace{-2mm}

Table~\ref{tab:adaptation_comparison_all} reports quantitative results under the unseen robot adaptation setting. Our method consistently outperforms V2V baselines across all metrics, especially in motion fidelity and embodiment correctness. Figure. \ref{fig:compare} further show that generic V2V models often suffer from motion drift, inconsistent limb articulation, and weak embodiment fidelity, whereas our adapted model better preserves motion trajectories while accurately reflecting the target robot embodiment.

\begin{table*}[!htbp]
\centering
\footnotesize
\setlength{\tabcolsep}{1.5pt}
\caption{Comparison of unseen robot adaptation on the Synthetic Held-out Embodiment Benchmark and Real-world Benchmark. Higher is better except MSE.}
\label{tab:adaptation_comparison_all}
\begin{tabular}{lccccccc|cccc}
\toprule
&
\multicolumn{7}{c|}{Synthetic Held-out Benchmark}
& \multicolumn{4}{c}{Real-world Benchmark} \\
\cmidrule(lr){2-8} \cmidrule(lr){9-12}
& PSNR $\uparrow$
& SSIM $\uparrow$
& MSE $\downarrow$
& Motion. $\uparrow$
& Embod. $\uparrow$
& BG. $\uparrow$
& Overall $\uparrow$
& Motion. $\uparrow$
& Embod. $\uparrow$
& BG. $\uparrow$
& Overall $\uparrow$ \\
\midrule
Kling O1
& 22.70 & 0.8951 & 0.0067 & 8.06 & 6.94 & 9.52 & 7.08
& 7.49 & 8.46 & 9.91 & \textbf{8.53} \\

Kling O3
& 22.76 & 0.8959 & 0.0065 & 8.76 & 7.90 & 9.32 & 7.42
& 7.47 & 8.34 & 9.82 & 8.21 \\

Runway Gen4
& 18.83 & 0.6575 & 0.0181 & 7.26 & 7.50 & 8.14 & 7.22
& 6.79 & 5.07 & 8.61 & 7.22 \\

Wan2.1-VACE
& 22.44 & 0.8599 & 0.0066 & 6.40 & 5.88 & 8.68 & 6.22
& 5.60 & 5.65 & 7.85 & 6.45 \\

X-Humanoid
& 23.03 & 0.8891 & 0.0057 & 8.94 & 8.04 & 9.78 & 7.53
& -- & -- & -- & -- \\

\midrule
Ours
& \textbf{25.47} & \textbf{0.9039} & \textbf{0.0033} & \textbf{9.06} & \textbf{8.43} & \textbf{9.94} & \textbf{7.92}
& \textbf{8.47} & \textbf{8.56} & \textbf{9.95} & 8.39 \\
\bottomrule
\end{tabular}
\end{table*}

\begin{figure}[t] \centering \animategraphics[width=1.0\linewidth]{5}{section/ablation_fig/frame_}{00}{09} 
\vspace{-5mm}
\caption{
\textbf{Effectiveness of Decoupled Attention.}
We compare our full model with a baseline without decoupled attention. 
As highlighted by red circles, the baseline suffers from rendering errors (``Wrong Details'') and physical inconsistencies (``Wrong Motion''), while our full model produces faithful embodiments and accurate motion transfer for both seen and unseen robot identities. 
Readers can click and play the clips using {\color{red}\textbf{Adobe Acrobat}}.
}
\label{fig:ablation} \end{figure}

\begin{table*}[!htbp]
\centering
\footnotesize
\setlength{\tabcolsep}{3pt}
\caption{Ablation study on OmniHumanoid variants across decoupling and distillation stages. Best results are in bold.}
\label{tab:ablation_study}
\begin{tabular}{lcccccccc}
\toprule
Method &
FPS $\uparrow$ &
PSNR $\uparrow$ &
SSIM $\uparrow$ &
MSE $\downarrow$ &
Motion. $\uparrow$ &
Embod. $\uparrow$ &
BG. $\uparrow$ &
Overall $\uparrow$ \\
\midrule
Teacher Model w/o Decoupling
& 0.21 & 21.21 & 0.8928 & 0.0098 & 6.35 & 2.53 & 8.56 & 6.80 \\
Teacher Model
& 0.10 & \textbf{25.47} & \textbf{0.9039} & \textbf{0.0033} & \textbf{9.06} & \textbf{8.43} & \textbf{9.94} & \textbf{7.92} \\
Causal Student
& 4.96 & 23.35 & 0.8841 & 0.0058 & 8.82 & 8.07 & 9.58 & 7.60 \\
Full Streaming Student
& \textbf{4.96} & 23.34 & 0.8900 & 0.0053 & 8.90 & 8.09 & 9.64 & 7.73 \\
\bottomrule
\end{tabular}
\end{table*}

\subsection{Ablation Studies}
\vspace{-2mm}

We conduct system-level ablation studies on the core design of OmniHumanoid. Since our architecture combines several coupled components, including branch-isolated attention, rolling embodiment-LoRA loading, two-stage training, and streaming distillation, some low-level components cannot be removed in isolation without breaking the training or adaptation pipeline. We therefore focus on two main factors: motion-appearance decoupling and streaming distillation. As shown in Table~\ref{tab:ablation_study}, we compare four variants: a teacher model without motion-appearance decoupling, the full bidirectional teacher, a causal student trained without self-forcing, and the full streaming student.

The results show that motion-appearance decoupling is the most critical component. Without decoupling, the embodiment score drops from 8.43 to 2.53, while the motion score decreases from 9.06 to 6.35. This indicates that directly mixing source-motion conditioning with target-embodiment rendering causes strong interference, making it difficult to preserve both transferable motion and robot appearance. In contrast, the bidirectional teacher with our decoupled design achieves the best reconstruction quality and the strongest motion, embodiment, and background consistency, validating the importance of explicitly separating transferable motion from embodiment-specific appearance.

Streaming distillation mainly improves efficiency at the cost of some generation quality. 
Compared with the bidirectional teacher, the causal student increases the speed from 0.10 FPS to 4.85 FPS, but its reconstruction and consistency metrics decrease. 
The full streaming student further improves efficiency to 4.96 FPS and slightly recovers quality compared with the causal student, improving SSIM, MSE, motion, embodiment, background, and overall scores. 
These results show that streaming distillation provides a practical speed-quality trade-off, while motion-appearance decoupling remains the key factor for stable cross-embodiment generation.

%% file: section/conclusion.tex
\section{Conclusion}
\vspace{-2mm}

We present a scalable framework for human-to-robot video generation that explicitly decouples transferable motion learning from robot-specific appearance modeling. By combining paired pretraining on a limited set of robot embodiments with lightweight, paired-free adaptation via RobotAppearance LoRA modules, our approach enables efficient generalization to unseen robots without retraining the core model. A decoupled attention architecture and rolling LoRA training strategy ensure minimal interference between motion transfer and appearance adaptation. Together with a large-scale synthetic dataset constructed through motion-aligned combinatorial rendering, our method reduces data collection cost while delivering strong motion fidelity and embodiment consistency, offering a practical path toward scalable robot data generation for embodied intelligence.